\documentclass[runningheads]{llncs}
\usepackage{marvosym}
\usepackage{graphicx}
\usepackage{makecell}
\usepackage{siunitx}

\usepackage{booktabs}
\usepackage{multirow}
\usepackage{tabularx}
\usepackage{array}

\begin{document}

\title{HierCAD: Hierarchical Text-to-CAD Design via Structure Alignment and Parameter Grounding}

\titlerunning{HierCAD}

\author{Jimin Xu, Tianbao Wang, Tao Jin, \and Zhou Zhao\textsuperscript{(\Letter)}}
\authorrunning{J. Xu et al.}
\institute{College of Computer Science and Technology, Zhejiang University, Hangzhou, China\\
\email{zhaozhou@zju.edu.cn}}

\maketitle              % typeset the header of the contribution

\begin{abstract}
Recent text-to-CAD approaches have shown promising results by leveraging large language models, but they often struggle with maintaining structural consistency in complex designs and accurately grounding geometric parameters. To address these issues, we propose HierCAD, a hierarchical text-to-CAD framework that improves both structural reasoning and parameter prediction. HierCAD reformulates CAD generation as progressive reasoning by decomposing CAD construction trees into object-level procedural reasoning and part-level topology reasoning trajectories. To further improve generation fidelity, we introduce a unified Structure Alignment and Parameter Grounding (SAPG) learning strategy. Structure alignment aligns topology reasoning trajectories with their corresponding parametric CAD spans, while parameter grounding mitigates shortcut learning through structure-preserving parameter perturbations and ranking-based supervision.
Experiments demonstrate that HierCAD outperforms prior state-of-the-art methods on both CAD sequence generation and reconstructed CAD model evaluation. Our code is available at \url{https://github.com/Collab-Gen/HierCAD}.

\keywords{Text-to-CAD  \and Large Language Models.}
\end{abstract}

\section{Introduction}

Computer-Aided Design (CAD) plays a crucial role in industrial design, mechanical engineering, and digital manufacturing~\cite{EN20130103}. 
Recent text-to-CAD approaches have shown promising results by leveraging large language models. Text2CAD~\cite{khan2024text2cad} introduces a large-scale annotation pipeline and a transformer architecture that predicts vectorized CAD sequences from textual prompts. CADmium~\cite{govindarajan2026cadmium} further demonstrates that strong code LLMs~\cite{hui2024qwen2} can be fine-tuned to directly generate JSON-formatted CAD sequences in a text-to-text fashion, significantly narrowing the gap between natural language understanding and sequential CAD design. Together, these works establish text-to-CAD as a promising research direction and show that modern generative models can recover substantial portions of CAD construction histories from language alone. However, despite these advances, accurately generating complex CAD objects from text remains far from solved.

Our analysis of existing methods reveals two major bottlenecks. First, current language model based approaches usually formulate CAD generation as flat autoregressive sequence prediction. This representation entangles heterogeneous information, including object composition, loop topology, and continuous geometric parameters, inside a single token stream. As a result, long-horizon procedural dependencies are difficult to maintain, especially for multi-part objects and sketches with many loops. 
Second, even when the coarse topology is correct, parameter prediction remains fragile. We observe that autoregressive models often rely on shortcut statistics and spuriously reuse previously generated legal values, leading to inter-field parameter confusion and intra-field parameter collapse. These errors degrade geometric fidelity and mesh quality even when the symbolic structure of the CAD sequences is largely preserved.

To address these challenges, we propose \textbf{HierCAD}, a hierarchical text-to-CAD framework that improves both structural reasoning and parameter grounding. We first \emph{reformulate hierarchical supervision}. Instead of learning a flat CAD sequence directly, we disentangle the CAD construction tree into multiple reasoning layers and supervise generation through two explicit trajectories: an object-level global procedural reasoning over parts and Boolean operations, and a part-level local topology reasoning over loop primitives. This reformulation turns CAD generation into progressive reasoning, allowing the model to resolve compositional structure before committing to geometric parameters. We then propose a unified \emph{structure alignment and parameter grounding (SAPG)} learning strategy. On the structural side, we align topology reasoning trajectories with their corresponding parametric CAD spans in representation space, encouraging the final CAD sequence to preserve the predicted loop structure. On the parameter side, we explicitly target shortcut learning using structure-preserving perturbations and ranking-based supervision, teaching the model to prefer text-grounded parameter assignments over spuriously plausible alternatives.
Qualitative and quantitative experiments show that HierCAD significantly
outperforms state-of-the-art method and better preserves part composition, loop structure, and parameter fidelity on long-horizon and multi-loop examples.
Our contributions are summarized as follows:
\begin{itemize}
    \item We propose a hierarchical reformulation of text-to-CAD generation that decomposes flat supervision into global procedural reasoning and local topology reasoning, providing a stronger structural prior for complex CAD generation.
    \item We introduce a learning strategy that combines structure alignment learning with parameter grounding learning, explicitly addressing topology drift and shortcut-driven parameter errors in autoregressive CAD generation.
    \item Qualitative and quantitative experiments show that HierCAD significantly outperforms state-of-the-art method on both generated CAD sequences and reconstructed CAD models.
\end{itemize}

\section{Related Work}
\label{sec:related_work}

\noindent\textbf{CAD datasets and procedural representations.}
Fusion360 \cite{willis2021fusion} provides a dataset with real user-authored CAD construction histories, and DeepCAD \cite{wu2021deepcad} introduces a large-scale CAD construction sequences dataset by parsing a large subset of ABC \cite{koch2019abc}, making it a common benchmark for CAD generation.
CAD-as-language \cite{ganin2021cadlang}, SketchGen \cite{para2021sketchgen}, Engineering Sketch Generation \cite{willis2021engineering}, and Vitruvion \cite{seff2022vitruvion} model CAD records as structured token, graph, or constraint processes, while SkexGen \cite{xu2022skexgen} and Hierarchical Neural Coding \cite{xu2023hnc} separate topology, geometry, extrusion, or controllable hierarchy.

\noindent\textbf{CAD generation and reconstruction.}
Direct approaches model meshes or B-reps with explicit geometric and topological organization \cite{nash2020polygen,lambourne2021brepnet,jayaraman2023solidgen,xu2024brepgen}. Reconstruction methods infer editable operations or programs from observations, including B-reps, multimodal CAD data, extrusion cylinders, and point clouds \cite{dupont2022cadopsnet,ma2023multicad,uy2022point2cyl,khan2024cadsignet,liu2024point2cad,li2024sfmcad,ma2024drawcad,dupont2024transcad}. Image- and sketch-conditioned systems also emphasize executable commands, consistent topology, and reliable parameters \cite{you2025img2cad,alam2025gencad,li2022free2cad}.

\noindent\textbf{Language-conditioned CAD design.}
Text2CAD \cite{khan2024text2cad} make CAD generation controllable from natural language. It introduces beginner-to-expert text supervision over DeepCAD \cite{wu2021deepcad} and trains a transformer to map prompts to CAD construction sequences, while CADmium \cite{govindarajan2026cadmium} shows that code LMs can generate minimal JSON CAD histories in a pure text-to-text setting. Recent LLM-based systems \cite{xu2024cadmllm,wang2025cadfusion,liao2025cadllm,zhang2025flexcad,guan2025cadcoder,alrashedy2024cadcode,yuan2024cadtalk,jones2025aidl} broaden the interface from fixed CAD datasets to language-guided design, but they also show that controllability depends on how the target program is represented and supervised. 

\section{Preliminary}

Recent text-to-CAD methods are largely built upon CAD sequences introduced by DeepCAD~\cite{wu2021deepcad}. Instead of representing a shape as meshes or point clouds, DeepCAD models a CAD object as an ordered sequence of sketching and modeling operations that preserve the construction history of the design. Each operation is parameterized by geometric attributes, such as sketch primitives, coordinates, radii, extrusion distances, and Boolean operations. CAD sequences preserve the construction history of the design, enabling exact geometry reconstruction and downstream CAD editing.

CADmium~\cite{govindarajan2026cadmium} further reformulates text-to-CAD generation as a text-to-text problem. Specifically, it performs supervised fine‑tuning on Qwen 2.5-Coder-14B~\cite{hui2024qwen2}, an instruction-tuned code LLM, to autoregressively generate JSON-formatted CAD sequences from natural language prompts. By serializing JSON into token streams, CADmium enables large language models to directly generate CAD sequences using standard next-token prediction, without relying on specialized vectorized representations or custom embedding layers.
The generation objective is to model:
\begin{equation}
    p_{\theta}(Y \mid X),
\end{equation}
where $X$ denote a natural-language CAD description and $Y$ denote the corresponding JSON-formatted CAD sequences.

\section{Methods}

\subsection{Hierarchical CAD Reasoning}

\begin{figure*}[t]
  \includegraphics[width=\linewidth]{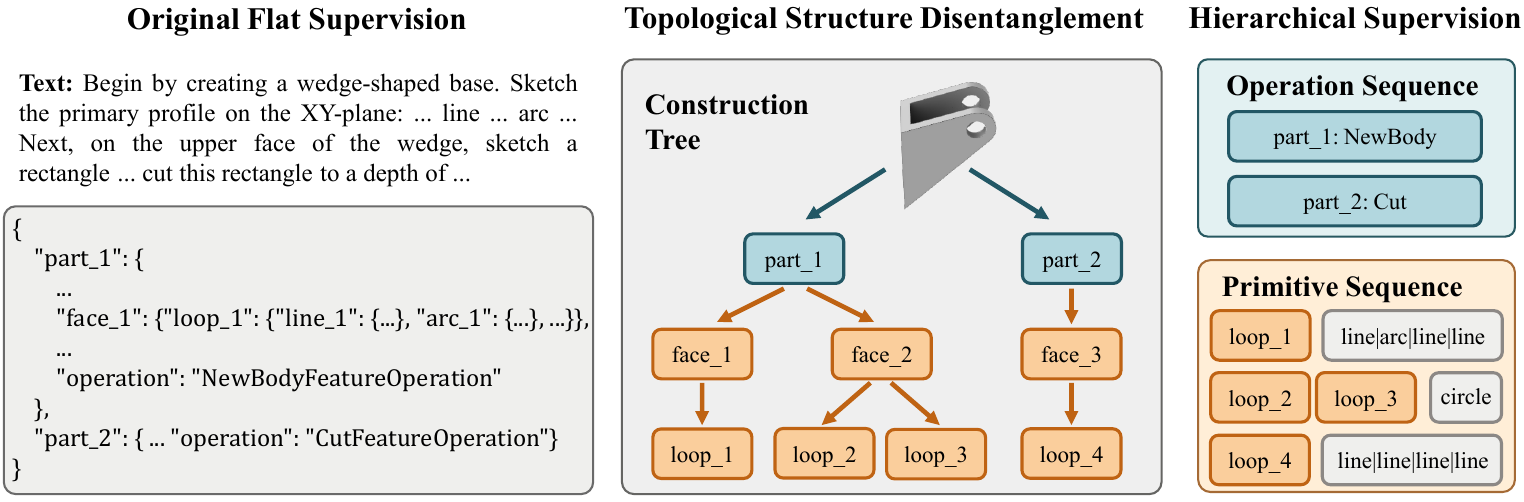}
  \caption {\textbf{Overview of hierarchical supervision.} HierCAD reformulates text-to-CAD generation with two levels of supervision: global procedural reasoning over part operation sequences, and local topology reasoning over loop primitive sequences.}
  \label{alg-data}
  \vspace{-2mm}
\end{figure*}

\noindent\textbf{Disentangle CAD Construction Tree.}
CAD sequences inherently entangle heterogeneous information, including topological structures and numerical parameters. Formulating CAD sequences as flat targets in language models significantly weakens the model’s ability to perform long-horizon procedural reasoning, which becomes particularly severe for long-tail multi-part CAD objects. To address this issue, we disentangle the CAD construction tree from the original CAD sequence. Instead of treating CAD generation as flat sequence modeling, we explicitly separate the CAD sequences into multiple reasoning layers, progressively reducing structural ambiguity before geometric parameter generation.

Formally, given a CAD sequence $\mathcal{C}$, we disentangle its topological structures from continuous geometric parameters by abstracting away numerical attributes while preserving the discrete hierarchical organization over parts, faces, and loops. Based on this decomposition, we derive the construction tree $T$ from $\mathcal{C}$. As shown in Figure~\ref{alg-data}, $T$ is organized into multiple structural levels:
\begin{equation}
T=\{P, F, L\},   
\end{equation}
where $P = \{p_i\}$ denotes object-level composition nodes, $F = \{f_j\}$ denotes part-level sketch faces and $L = \{l_k\}$ denotes closed loops.

\noindent\textbf{Reformulate Hierarchical Supervision.}
Based on the disentangled construction tree $T$, we further reformulate the CAD sequence into a hierarchical supervision target. Inspired by the effectiveness of Chain-of-Thought (CoT)~\cite{wei2022chain} supervision in complex reasoning tasks, we construct hierarchical CoT annotations aligned with the CAD construction tree. Specifically, we decompose the supervision into two levels: object-level global procedural reasoning and part-level local topology reasoning.

At the object level, we linearize the parts $P=\{p_i\}_{i=1}^{N}$ in composition order to obtain a global supervision
\begin{equation}
    \mathcal{G} = \big[(p_1, o_1), (p_2, o_2), \dots, (p_N, o_N)\big],
\end{equation}
where $o_i \in \mathcal{O}$ is the Boolean operation attached to part $p_i$. It contains four categories:
\begin{equation}
    \mathcal{O} = \{\texttt{New}, \texttt{Join}, \texttt{Cut}, \texttt{Intersect}\}.
\end{equation}
Intuitively, $\mathcal{G}$ acts as a coarse-grained procedural reasoning trajectory that explicitly describes how the final object should be composed step-by-step. By exposing this reasoning before the final parameterized CAD sequence is generated, it strengthens long-horizon reasoning. This is particularly important for long-tail multi-part CAD objects, where explicit procedural planning significantly reduces autoregressive ambiguity.

For each part node $p_i$, we derive a local supervision target $\mathcal{P}_i$ by linearizing its loop nodes in topological order:
\begin{equation}
    \mathcal{P}_i = \big[(l_{i,1}, q_{i,1}), (l_{i,2}, q_{i,2}), \dots, (l_{i,M_i}, q_{i,M_i})\big],
\end{equation}
where $l_{i,j}$ denotes the $j$-th loop in part $p_i$, and $q_{i,j}$ denotes the closed primitive sequence associated with that loop. The primitive vocabulary contains three categories:
\begin{equation}
    \mathcal{V}_{prim} = \{\texttt{line}, \texttt{arc}, \texttt{circle}\}.
\end{equation}
Each $q_{i,j}$ is an ordered closed-loop sequence formed by one or more primitives from $\mathcal{V}_{prim}$, such as $\texttt{line}| \texttt{arc}| \texttt{line}| \texttt{line}$ or $\texttt{circle}$ shown in Figure~\ref{alg-data}. Intuitively, $\mathcal{P}_i$ provides a fine-grained topology reasoning trajectory for part $p_i$: instead of directly predicting coordinates and extrusion parameters, the model first determines which loops should exist in the part and how each loop should be sketched as a closed primitive sequence. Such topology-aware supervision provides strong structural priors during generation, improving sketch validity and topological coherence.

Overall, our hierarchical supervision reformulates CAD generation as progressive reasoning:
\begin{equation}
P(Y\mid X)
=
P(\mathcal{G}\mid X)
\prod_{i=1}^{N}
P(\mathcal{P}_i \mid \mathcal{G}, X),    
\end{equation}
where $X$ denotes the textual description, $\mathcal{G}$ denotes the global procedural reasoning sequence and $\mathcal{P}_i$ denotes the local topology-aware generation process for the $i$-th CAD part. This hierarchical formulation enables the model to progressively resolve procedural dependencies and geometric structures before parameter instantiation, substantially improving structural consistency and compositional generalization for complex CAD generation.

\subsection{Structure Alignment Learning}

We optimize the language model using autoregressive teacher-forcing loss under hierarchical supervision:
\begin{equation}
    \mathcal{L}_{ce}
    =
    - \sum_{t \in \mathcal{T}_{sup}}
    \log p_{\theta}(y_t \mid X, y_{<t}),
\end{equation}
where $\mathcal{T}_{sup}$ denotes the supervised assistant region. 

While this objective ensures correct serialization of hierarchical reasoning trajectories and parametric CAD sequences, it does not explicitly enforce structure alignment between them. For example, the topology reasoning trajectory may correctly predict the sketch structure, whereas the generated parametric CAD sequence fails to preserve it due to accumulated autoregressive errors. This issue becomes particularly severe for long-horizon multi-loop sketches, where structural dependencies are difficult to maintain through token-level supervision.

To address this issue, We introduce structure alignment learning that aligns topology reasoning trajectory with their corresponding parametric CAD sequences. For loop $l_k$, we identify the aligned token spans and obtain pooled representations $h^{r}_{k}$ and $h^{s}_{k}$ by mean aggregation over the final-layer hidden states of the decoder.
Since both representations describe the same underlying CAD structure, they should remain close in the representation space. We therefore define the alignment loss as: 
\begin{equation}
    \mathcal{L}_{struct}
    =
    \frac{1}{|\mathcal{S}|}
    \sum_{k \in \mathcal{S}} (1 - \cos\left(h^{r}_{k}, h^{s}_{k}\right)),
\end{equation}
where $\mathcal{S}$ denotes the set of loops sampled. By aligning topology reasoning with parametric sequences, this objective strengthens topological coherence for complex multi-loop CAD generation.

\subsection{Parameter Grounding Learning}
\label{pgl}

While hierarchical reasoning and structure alignment learning substantially improve topology coherence, numerical parameter prediction still suffers from shortcut learning behaviors. 
We attribute this issue to spurious parameter correlations arising from superficial numerical co-occurrence in training data, where unrelated parameter slots may coincidentally share identical values. Consequently, autoregressive language models tend to exploit previously generated legal values as shortcut signals rather than learning long-range textual description grounding.

Specifically, we observe two dominant shortcut-induced failure patterns. The first is inter-field parameter confusion, where numerical values are spuriously reused across semantically unrelated parameter slots, such as confusing pose or extrusion parameters with point coordinates. The second is intra-field parameter collapse, where distinct slots within the same parameter field are compressed to a smaller set of repeated values, such as assigning multiple loop endpoints or circle centers to identical coordinates. 

Based on this observation, we construct structure-preserving parameter perturbations that maintain plausible CAD structures while making the numerical parameters inconsistent with the textual description. For each training example, we construct a perturbed target:
\begin{equation}
y^{-} = q_{\phi}(y),
\end{equation}
where $q_{\phi}(\cdot)$ denotes parameter perturbation operators. Specifically, for a loop with geometric center $c$, each point $p$ is transformed as:
\begin{equation}
p' = c + A(p-c) + b,
\end{equation}
where $A$ denotes a local anisotropic scaling matrix and b denotes a small translation offset. This transformation jointly perturbs line endpoints, arc control points, and circle centers while maintaining local connectivity and overall loop structure.

Let $\mathcal{T}_{par}$ denote the set of parameter tokens in the original target. We first apply parameter-specific supervision:
\begin{equation}
    \mathcal{L}^{+}
    =
    - \frac{1}{|\mathcal{T}_{par}|}
    \sum_{t \in \mathcal{T}_{par}}
    \log p_{\theta}(y_{t} \mid X, y_{<t}).
\end{equation}

Let $\mathcal{T}^{-}_{par}$ denote the parameter-token set in the perturbed target $y^{-}$. We further compute the perturbed parameter loss:
\begin{equation}
    \mathcal{L}^{-}
    =
    - \frac{1}{|\mathcal{T}^{-}_{par}|}
    \sum_{t \in \mathcal{T}^{-}_{par}}
    \log p_{\theta}(y^{-}_{t} \mid X, y^{-}_{<t}).
\end{equation}

We then encourage the model to assign a higher likelihood to the original target than to the perturbed counterpart via:
\begin{equation}
    \mathcal{L}_{param}
    =
    \max \bigl(0,\; m + \mathcal{L}^{+} - \mathcal{L}^{-}\bigr),
\end{equation}
where $m$ denotes a ranking margin. This objective encourages the model to ground parameter prediction on long-range textual descriptions rather than exploiting previously generated legal values.

\section{Experiments}

\subsection{Experimental Setup}
\noindent\textbf{Datasets.}
We conduct experiments using the DeepCAD dataset and annotations from Text2CAD and CADmium. Specifically, we use the Minimal JSON format released by Text2CAD, which removes random keys and redundant information from the original DeepCAD JSON. For textual descriptions, we use the expert-level descriptions re-annotated by CADmium, which are more human-like and concise. On top of these annotations, we construct our hierarchical supervision targets by reformulating each JSON-formatted CAD sequence into object-level and part-level reasoning scaffolds. 

\noindent\textbf{Implementation Details.}
Our models are built on top of Qwen 2.5-Coder-14B and fine-tuned with LoRA applied to all linear projections in the transformer blocks. The training configuration uses LoRA rank $r=64$, scaling factor $\alpha=16$, and dropout $0.1$. 
We optimize the model with AdamW, using a peak learning rate of $2\times 10^{-4}$, cosine decay, 100 warmup steps, weight decay of $10^{-3}$.
The model is trained for approximately 29,000 steps using loss $\mathcal{L}_{ce}$ and $\mathcal{L}_{struct}$ weighted by 1.0 and 0.2, respectively. At the last 12,000 steps, $\mathcal{L}_{param}$ is applied using a weight of 0.2 and a margin of 1.0. The complete training process takes approximately 80 hours on four NVIDIA RTX A6000 GPUs.

\noindent\textbf{Baselines.}
We compare HierCAD against two prior text-to-CAD methods: Text2CAD is built on a vectorized CAD representation and a specialized transformer architecture. CADmium fine-tunes Qwen 2.5-Coder-14B and directly generates flat JSON-formatted CAD sequences.
In addition, we report an ablation denoted as HierCAD w/o SAPG. This variant replaces the flat target in CADmium with our hierarchical supervision target. It does not include the structure alignment and parameter grounding learning mechanisms introduced in the full HierCAD. This ablation allows us to quantify how much of the performance gain comes purely from the dataset and target reformulation itself.

\begin{table*}[t]
\caption{Quantitative evaluation. Best performances are in bold.}
\label{tab1}
\centering
\renewcommand{\arraystretch}{1.2}
\begin{tabularx}{\linewidth}{
l|
>{\centering\arraybackslash}X|
>{\centering\arraybackslash}X|
>{\centering\arraybackslash}X|
>{\centering\arraybackslash}X
}

\toprule
\multirow{2}{*}{Metric}
& \multirow{2}{*}{Text2CAD}
& \multirow{2}{*}{CADmium}
& HierCAD
& HierCAD \\

& & & w/o SAPG & Full \\
\midrule

Line F1 $\uparrow$
& 81.18
& 91.25
& 92.16
& \textbf{93.12} \\

Arc F1 $\uparrow$
& 36.27
& 75.68
& 75.87
& \textbf{79.43} \\

Circle F1 $\uparrow$
& 75.74
& 91.00
& 93.26
& \textbf{93.91} \\

Extrusion F1 $\uparrow$
& 94.88
& 98.21
& 99.09
& \textbf{99.10} \\

IR (\%) $\downarrow$
& 3.40
& 4.12
& 1.96
& \textbf{1.45} \\

CD mean $\downarrow$
& 63.47 
& 44.52 
& 42.17 
& \textbf{35.22} \\

CD median $\downarrow$
& 0.54
& 0.24
& 0.23
& \textbf{0.21} \\

SIR $\downarrow$
& 0.05
& 0.05
& 0.04
& \textbf{0.04} \\

DangEL $\downarrow$
& 1.25
& 1.90
& 1.17
& \textbf{1.10} \\

SegE $\downarrow$
& 0.58
& 0.87
& 0.45
& \textbf{0.39} \\

FluxEE ($\times 10^2$) $\downarrow$
& 0.61
& 0.46
& 0.41
& \textbf{0.40} \\

Watertightness (\%) $\uparrow$
& 87.24
& 87.87
& 88.24
& \textbf{89.79} \\

EECM $\uparrow$
& 0.83
& 0.89
& 0.91
& \textbf{0.93} \\

DMCD ($\times 10^3$) $\downarrow$
& 5.70
& 2.68
& 2.11
& \textbf{1.76} \\

SD mean ($\times 10^2$) $\downarrow$
& 7.15
& 2.04
& 1.20
& \textbf{0.99} \\

SD median ($\times 10^2$) $\downarrow$
& 1.30
& \textbf{0.00}
& \textbf{0.00}
& \textbf{0.00} \\

\bottomrule
\end{tabularx}
\vspace{-2mm}
\end{table*}

\subsection{Quantitative Evaluation}
Table~\ref{tab1} reports the quantitative evaluation.
Following Text2CAD and CADmium, we evaluate the models from two complementary perspectives: the quality of the generated CAD sequences and the geometric as well as topological quality of the reconstructed CAD models. Overall, the results show a consistent trend. Replacing flat supervision with our hierarchical target already yields clear gains over CADmium, and the full HierCAD further improves almost all reported metrics, confirming that structure alignment and shortcut-aware parameter refinement are both necessary for robust text-to-CAD generation.

\noindent\textbf{Evalution on generated CAD sequences.}
We evaluate the predicted CAD sequences using F1 scores. These metrics measure whether the generated construction sequence recovers the correct sketch primitives and extrusion operations by matching predicted and ground-truth loops. HierCAD achieves the best performance across all four sequence metrics, including Line, Arc, Circle, and Extrusion. Compared with CADmium, the most substantial gain appears on Arc F1, indicating that the proposed hierarchical reasoning and structure alignment are especially helpful for preserving complex multi-primitive loop structures.

\noindent\textbf{Evalution on reconstructed CAD models.}
We evaluate the reconstructed CAD models using both geometric and topological metrics adopted from CADmium. HierCAD achieves the lowest invalidity ratio at 1.45\%. It also obtains the best Chamfer Distance, which indicates more accurate geometric reconstruction after converting the generated sequence into a CAD model. Beyond these coarse similarity measures, HierCAD also performs best in Watertightness, EECM, and DMCD, showing that the predicted models better preserve manifold validity, topological invariants, and local surface geometry. The improvements in Sphericity Discrepancy (SD) further indicate that our method better matches the global compactness of the target shapes. Taken together, these results show that HierCAD improves not only symbolic CAD sequence accuracy, but also the final manufacturable quality of reconstructed 3D objects.

\begin{figure*}[t]
  \includegraphics[width=\linewidth]{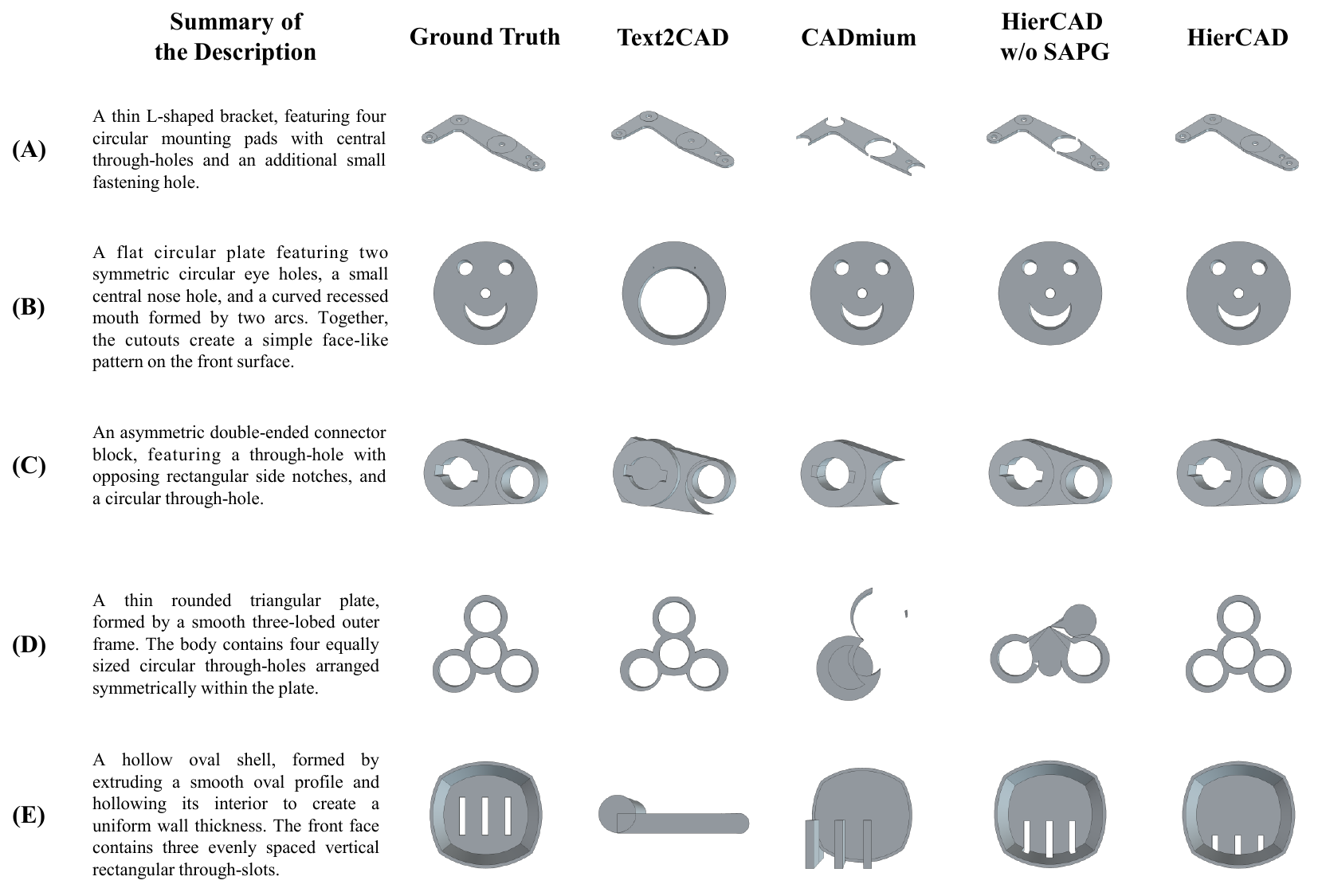}
  \caption {Qualitative comparison on reconstructed CAD models.}
  \label{exp-main}
  \vspace{-2mm}
\end{figure*}

\begin{figure*}[t]
  \includegraphics[width=\linewidth]{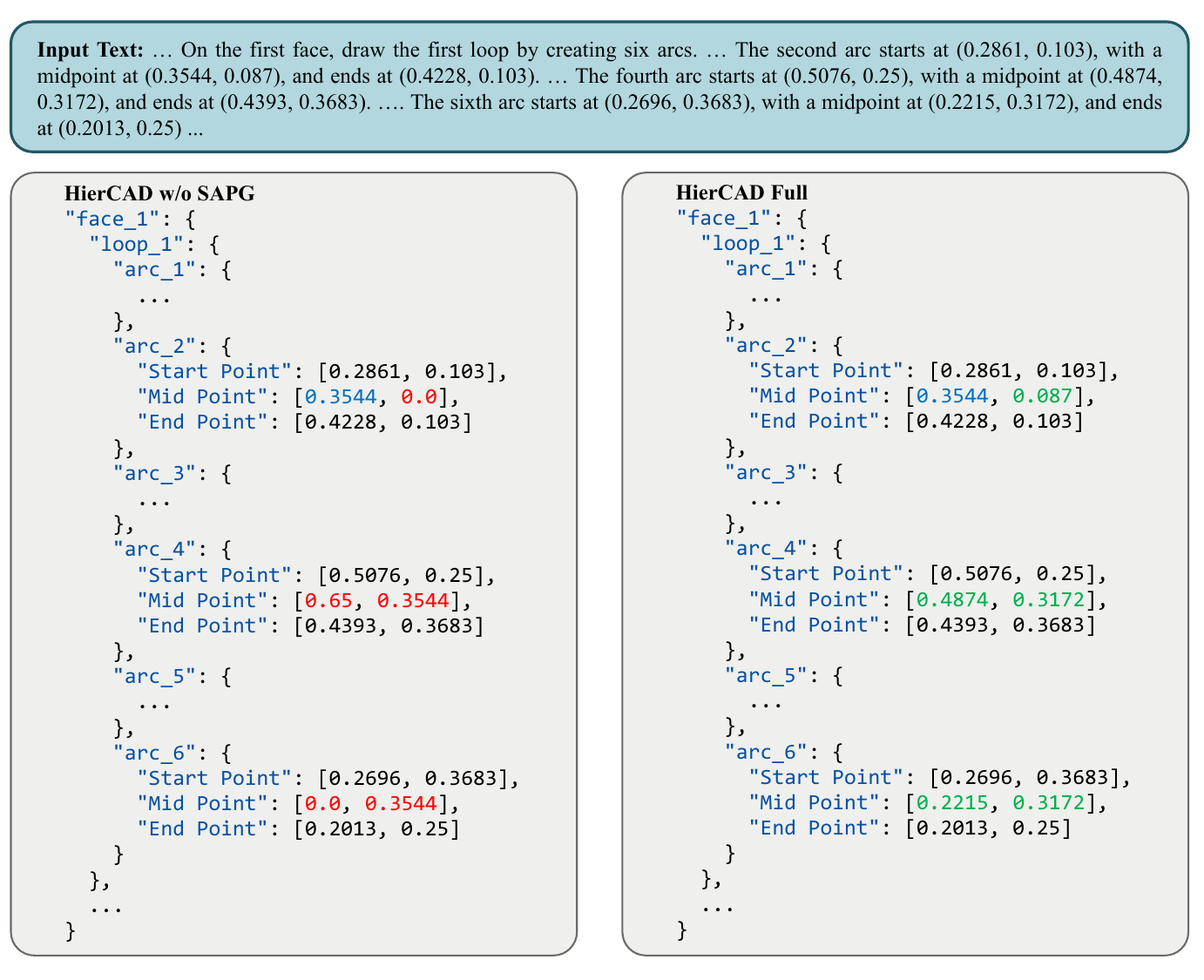}
  \caption {Comparison on generated CAD sequences.}
  \label{exp-2}
  \vspace{-2mm}
\end{figure*}

\subsection{Qualitative Comparison}
We showcase five representative test cases in Figure~\ref{exp-main} to qualitatively compare the generated CAD models and better understand behavioral differences among methods. Overall, we observe a clear progression in error patterns. Text2CAD most frequently fails at the level of construction decomposition itself, including missing parts, hallucinated extra parts, and collapsing multi-primitive loops into over-simplified circle-based structures. CADmium is typically closer to the target than Text2CAD, but still often breaks the intended constructive semantics, for example by changing Boolean operations, reorganizing face-loop assignments, or drifting substantially in sketch geometry despite preserving part of the loop inventory. HierCAD w/o SAPG already recovers much more of the intended topology, indicating that the hierarchical target reformulation alone provides a strong structural prior. The full HierCAD further improves both topology preservation and parameter fidelity, producing the most stable predictions across all selected cases.

Specifically, in Figure~\ref{exp-main} (B), Text2CAD collapses a two-arc loop into a circle. In Figure~\ref{exp-main} (C), Text2CAD predicts an unnecessary additional joined part, and CADmium over-generates loops while rewriting one target face into a different decomposition. In contrast, HierCAD recovers the exact structure. In addition, in Figure~\ref{exp-main} (E), Text2CAD fails to generate the intended construction program. CADmium incorrectly predicts \texttt{Cut} operations as \texttt{NewBody}. Although HierCAD exhibits slight pose drift in this case, it still faithfully recovers the exact part-level topological structure and Boolean operation sequences. These cases demonstrate that the hierarchical supervision substantially improves structural reasoning and reduces constructive hallucination.

Figure~\ref{exp-main} (A) presents a particularly challenging long-horizon single-part example with six faces and eleven loops. Here, CADmium and HierCAD w/o SAPG both lose part of the target loop structure, whereas the full HierCAD preserves the complete face-loop organization and exactly matches the target. This case illustrates the purpose of the proposed structure alignment learning: the key challenge is not merely to reason about the correct loops, but to consistently maintain that topology in the final parametric CAD sequence.

Figure~\ref{exp-main} (D) reveals the benefit of parameter grounding once the coarse structure is already correct: all four methods recover the same one-face, five-loop sketch topology, but they differ noticeably in geometric fidelity. Text2CAD and HierCAD w/o SAPG produce mild arc drift, while CADmium exhibits a larger scale shrinkage and more visible sketch deformation. The full HierCAD removes these residual deviations and exactly recovers the target parameters.

\subsection{Ablation Studies}
To better understand the effect of Structure Alignment and Parameter Grounding (SAPG) Learning, we present a representative example in Figure~\ref{exp-2}. The input text describes a rounded triangular frame composed of six connected arcs and four circular holes. While the overall topology is relatively simple, accurate reconstruction requires grounding multiple arc midpoint coordinates from different parts of the input text.

The ablated model produces numerically plausible but semantically incorrect values. Instead of grounding each midpoint from the corresponding geometric description, it repeatedly copies values that appear elsewhere in the sequence. For example, the value $0.3544$ originally corresponds to the x-coordinate of multiple circle centers, yet it is spuriously reused as the y-coordinate of both \texttt{arc\_4.mid} and \texttt{arc\_6.mid}. Similarly, the boundary scale value $0.65$ is incorrectly copied into the midpoint coordinate of \texttt{arc\_4}, while several midpoint coordinates collapse to the frequently occurring value $0.0$.

These errors are consistent with the shortcut-learning phenomenon discussed in Section~\ref{pgl}. Rather than grounding parameter prediction on long-range textual descriptions, the autoregressive model exploits previously generated legal numerical values as shortcut signals. Consequently, parameters from unrelated semantic fields (e.g., circle centers, sketch scale, and arc control points) become confused, producing inter-field parameter confusion. Moreover, multiple midpoint coordinates are compressed into a small set of repeated values (e.g., $0.0$ and $0.3544$), exhibiting intra-field parameter collapse.

In contrast, HierCAD with SAPG recovers all midpoint coordinates exactly and matches the ground-truth CAD sequence. This example demonstrates that SAPG effectively discourages shortcut-based numerical copying and encourages grounding parameter prediction in the corresponding textual evidence, leading to substantially more faithful geometric reconstruction.

\section{Conclusion}

We present HierCAD, a hierarchical text-to-CAD framework that addresses two core limitations of existing autoregressive CAD generators: weak long-horizon structural reasoning under flat sequence supervision, and shortcut-driven errors in continuous parameter prediction. Extensive experiments demonstrate that HierCAD consistently improves both generated CAD sequences and reconstructed CAD models over baselines. These results suggest that progress in text-to-CAD generation depends not only on stronger foundation models, but also on better supervision reformulation and training objectives tailored to CAD construction processes. Overall, HierCAD provides an effective path towards more accurate, valid, and editable text-driven CAD generation. We hope that this framework can serve as a useful step toward more reliable CAD copilots for engineering and design applications.

\subsubsection*{Acknowledgements.}
This work was supported by the National Natural Science Foundation of China
under Grant No. U24A20326.

\bibliographystyle{splncs04}
\bibliography{custom}

\end{document}